# An Evolutionary Multitasking Algorithm with Multiple Filtering for High-Dimensional Feature Selection

Lingjie Li, Manlin Xuan, Qiuzhen Lin, *Member, IEEE*, Min Jiang, *Senior Member, IEEE*, Zhong Ming, and Kay Chen Tan, *Fellow, IEEE*

*Abstract*—Recently, evolutionary multitasking (EMT) has been successfully used in the field of high-dimensional classification. However, the generation of multiple tasks in the existing EMT-based feature selection (FS) methods is relatively simple, using only the *Relief-F* method to collect related features with similar importance into one task, which cannot provide more diversified tasks for knowledge transfer. Thus, this paper devises a new EMT algorithm for FS in high-dimensional classification, which first adopts different filtering methods to produce multiple tasks and then modifies a competitive swarm optimizer to efficiently solve these related tasks via knowledge transfer. First, a diversified multiple task generation method is designed based on multiple filtering methods, which generates several relevant low-dimensional FS tasks by eliminating irrelevant features. In this way, useful knowledge for solving simple and relevant tasks can be transferred to simplify and speed up the solution of the original high-dimensional FS task. Then, a competitive swarm optimizer is modified to simultaneously solve these relevant FS tasks by transferring useful knowledge among them. Numerous empirical results demonstrate that the proposed EMT-based FS method can obtain a better feature subset than several state-of-the-art FS methods on eighteen high-dimensional datasets.

*Index Terms*— Feature selection, evolutionary multitasking, competitive swarm optimizer, evolutionary algorithm, high-dimensional classification

## I. Introduction

WITH the development of big data technologies, many real-world problems often consist of a large number of features, which poses a tremendous challenge for learning algorithms [1]. In particular, there are many redundant or irrelevant features in these problems, which may deteriorate the performance of the learning algorithms. In this regard, feature selection (FS) methods play a crucial role in data preprocessing, with the aim of reducing the dimensionality of data by eliminating redundant and irrelevant features. Thus, FS methods can simplify the learning process and achieve good performance by selecting a small subset of relevant features for training [2]. In general, most existing FS methods can be classified into three main categories, i.e., filter-based [3]-[5], wrapper-based [6]-[8], and embedded-based [9]-[11]. Specifically, filter-based FS methods select features with high relevance based on their intrinsic properties measured using univariate statistics [3], wrapper-based FS methods measure the "usefulness" of features based on the classifier's performance [6], and embedded-based FS methods automatically conduct the FS task during the learning process of the classifier [9]. Generally, different kinds of FS methods exhibit distinct characteristics and advantages, e.g., filter-based FS methods often have a low computational cost, while the other two kinds of FS methods provide better classification accuracy. The different kinds of FS methods mentioned above are combined in [12]-[14], improving overall performance.

However, when running classification in high-dimensional datasets (i.e., high-dimensional classification), most of the FS methods mentioned above still face two main challenges [15]-[16]: high computational cost and stagnation in local optima. To address these two issues, evolutionary algorithms (EAs), such as the genetic algorithm (GA) [17], differential evolution (DE) [18], and particle swarm optimizer (PSO) [8], have been widely used in FS due to their excellent global search ability. In particular, PSO is mostly used in existing FS methods due to its high efficiency and simple implementation [19]. A brief review of EA-based FS methods is given in **Section II. D**. Although empirical results have shown the effectiveness of EA-based FS methods, most of them still suffer from slow convergence speed or even premature convergence when handling complex high-dimensional FS problems with a large number of features [16].

To address the above issues, the evolutionary multitasking (EMT) method has been proposed as a prominent approach for high-dimensional classification due to its efficient search capability and fast convergence speed in solving complex problems [20]. Generally, EMT-based FS methods construct some low-dimensional relevant tasks from the original high-dimensional FS task and then use EAs to simultaneously solve these tasks through knowledge transfer [13]. In this way, useful knowledge for solving simple and relevant tasks can be transferred to simplify and speed up the solution of the original high-dimensional FS task. Chen *et al.* demonstrated the first attempt to use EMT for high-dimensional classification [13], called PSO-EMT, which uses PSO to simultaneously optimize

Manuscript received xx. xx. 2022; revised xx. xx. 2022; accepted xx. xx. 2022. This work was supported by the National Natural Science Foundation of China (NSFC) under Grant U21A20512, Grant 61876162, and Grant 61876110; the Research Grants Council of the Hong Kong under Grant PolyU11209219; the Guangdong "Pearl River Talent Recruitment Program" under Grant 2019ZT08X603; and the Shenzhen Science and Technology Innovation Commission under Grant R2020A045 and Grant JCYJ20190808164211203. (Corresponding author: *Qiuzhen Lin*)

L.J. Li, M.L. Xuan, Q.Z. Lin and Z. Ming are affiliated with the College of Computer Science and Software Engineering, Shenzhen University, Shenzhen 518060, China (e-mail of Q.Z. Lin: qiuzhlin@szu.edu.cn).
M. Jiang is affiliated with the School of Informatics, Xiamen University, Xiamen 361005, China (e-mail: minjiang@xmu.edu.cn).
K.C. Tan is affiliated with the Department of Computing, The Hong Kong Polytechnic University, Hong Kong SAR (e-mail: kctan@polyu.edu.hk).

the relevant tasks. In this method, one FS task with a small subset of relevant features is constructed from the original FS task by using the *Relief-F* method [3]. Then, these two relevant tasks are solved simultaneously by using a PSO variant to perform knowledge transfer among them, thus accelerating the convergence speed of the solving process of the original FS task. Subsequently, to improve the performance of PSO-EMT, Chen *et al.* devised a new multiple task generation strategy [21], called MTPSO, which applies the *Relief-F* method to generate several relevant tasks. Notably, MTPSO considers not only relevant features but also partially irrelevant features when constructing tasks. In addition, MTPSO employs a PSO variant to perform knowledge transfer between tasks.

Although the empirical results in [13] and [21] validate the superior performance of EMT-based FS methods in terms of solution quality and convergence speed for high-dimensional FS problems, the performance of these methods still depends heavily on the adopted multiple task generation and optimization methods. Specifically, multiple task generation in MTPSO and PSO-EMT uses only a single filtering method to determine the importance and relevance of features, but these methods cannot construct diversified relevant tasks and may not be effective for processing various high-dimensional datasets. Moreover, considering the multiple task optimization method, PSO variants are employed to perform knowledge transfer and show limitations in convergence speed and search capability as the feature dimensionality increases [22]. Therefore, the multiple task generation and optimization methods used in these existing EMT-based FS methods deserve further study.

Based on the above analysis, this paper devises a more effective EMT-based FS method for high-dimensional classification, called MF-CSO, which improves multiple task generation using multiple filtering methods and multiple task optimization using a modified competitive swarm optimizer (CSO). In summary, the main contributions of this paper are as follows:

1) An improved multiple task generation strategy is designed. Several filtering methods are used to construct more diversified relevant tasks. In this way, the improved diversity of the relevant tasks can help to enhance the knowledge transfer performance of EMT when handling various kinds of high-dimensional datasets.
2) An efficient multiple task optimization strategy is proposed by modifying a CSO variant. In this way, solving the original FS task can be highly accelerated by transferring useful knowledge from other relevant FS tasks.
3) An improved EMT-based FS method is proposed based on the above two strategies, and its effectiveness is evaluated on eighteen real-world high-dimensional datasets with the number of features ranging from 2000 to 12000. Extensive experimental results demonstrate that our method can obtain a subset of features with higher quality than several state-of-the-art EA-based FS methods and some traditional filtering methods.

The remaining parts of this paper are organized as follows. **Section II** introduces background information and related works on high-dimensional classification. **Section III** provides the details of the proposed method, and **Section IV** provides the experimental results and analysis. Finally, **Section V** draws conclusions and anticipates future work.

## II. PRELIMINARIES

This section first introduces some background information about FS, EMT and CSO, then reviews the research on EA-based FS methods for high-dimensional classification, and finally clarifies our motivations.

### A. Feature Selection.

Feature selection is essentially a combinatorial optimization problem, as there are $2^N$ different subsets of features to choose from a dataset with $N$ features [15]. Assuming that a labeled dataset $D$ consists of data with $N$ features and $H$ samples, the main purpose of an FS problem is to select $n$ features ($n<N$) from the original feature set $D$ such that the given performance metric $f(\cdot)$ is optimal, where $f(\cdot)$ denotes the classification error rate. Therefore, an FS problem with $N$ features can be formulated as follows:

$$\begin{aligned}\min\ &f(X)\\ s.t.\quad &X=(x_1,x_2,...,x_N)\\ &x_i\in\{0,1\}, i=1,2,3...,N\end{aligned} \quad (1)$$

where $X$ indicates a solution with $N$ features, and $x_i=1$ means that the corresponding $i$-th feature will be selected; otherwise, $x_i=0$ means that it will not be selected.

### B. Evolutionary Multitasking.

The EMT method aims to promote the transfer of useful knowledge to assist in the efficient solving of multiple tasks with different properties [20] since solving one optimization task can be helpful for solving other relevant optimization tasks by sharing useful knowledge among them. Mathematically, a multiple task optimization framework consisting of $K$ tasks can be described as follows:

$$\begin{aligned}\min\ &f_i(X_i),\quad i=1,2,...,K\\ s.t.\quad &X_i=[x_i^1,x_i^2,...,x_i^{D_i}]\end{aligned} \quad (2)$$

where $f_i(X_i)$ represents the objective function of task $i$ with its solution $X_i=[x_i^1,x_i^2,...,x_i^{D_i}]$, and $D_i$ is the dimensionality of $X_i$. The main purpose of EMT is to find a set of optimal solutions simultaneously in a large search space with certain efficient knowledge transfer methods.

How to handle the unique search space for each task and how to perform knowledge transfer among tasks are two main components of EMT [23]. In particular, effective knowledge transfer is crucial for EMT to perform knowledge transfer based on a single individual (KTS) or multiple individuals (KTM) [24]. In MFEA [20], TWO-MFEA [26], MFEA-DRA [27], and MFEA-II [28], KTS is realized by assortative mating and vertical culture propagation [25] to conduct knowledge transfer, while in MFPSO [29], AMFPSO [30], and MTMSO [31], KTM uses PSO to perform knowledge transfer among several relevant tasks.

On the other hand, EMT has shown effectiveness in solving certain real-world optimization problems. Due to the interdependent relationship in most practical problems, the experience of solving one problem can be helpful to solving another relevant problem. For example, Zhou *et al.* proposed a permuta-tion-based EMT method for solving the vehicle route-planning (VRP) problem [32], where the optimal solutions of one VRP problem can be transferred to guide the search of another VRP problem in the routing space. Zhang *et al.* presented an effective surrogate-assisted EMT algorithm using genetic programming to solve the dynamic flexible job shop scheduling (DFJSS) problem [33], in which newly generated individuals are evaluated by the built surrogates and then transferred to suitable JSS tasks. Shen *et al.* extended the EMT framework to solve real-world network reconstruction problems [34], in which similar network structure patterns between different networks can be fully exploited by transferring useful knowledge among these relevant networks.

*C. Competitive Swarm Optimizer.*

*Cheng* and *Jin* [35] first proposed the competitive swarm optimizer (CSO) and introduced the concept of a competition mechanism between particles within a single swarm. Unlike traditional PSOs, which use the global best and personal best particles to guide the swarm search, CSO applies a pairwise competition to evolve the swarm, showing stronger exploration ability and faster convergence speed than PSO [36]. In CSO, the swarm is first divided into two groups via random pairwise competition, resulting in winner particles with better performance and loser particles with worse performance. Then, the velocity and position of loser particles are updated by learning from their corresponding winner particles. This process can be formulated as follows:

$$V_L(t+1) = r_1 \times V_L(t) + r_2 \times (X_W(t) - X_L(t)) \\ + \varphi \times r_3 \times (\bar{X}(t) - X_L(t)) \quad (3)$$

$$X_L(t+1) = X_L(t) + V_L(t+1) \quad (4)$$

where $r_1$, $r_2$, and $r_3$ represent three random numbers ranging from 0 to 1. $V_L(t)$ and $X_L(t)$ indicate the velocity and position of the loser particle in the current *t*-th generation, respectively. $X_W(t)$ is the position of the corresponding winner particle in the current generation. $V_L(t+1)$ and $X_L(t+1)$ denote the velocity and position of the loser particle in the next generation, respectively. $\bar{X}(t)$ represents the mean position value of particles in the current swarm [35], and $\varphi$ is set to control the influence of $\bar{X}(t)$. Empirical results have verified the effectiveness of CSO in solving problems with large-scale variables [36]-[37]. This paper can be considered the first extension of CSO to transfer knowledge between related tasks.

*D. Related Works and Motivations.*

To date, many EAs have been successfully applied in the field of high-dimensional classification due to their efficient global search ability. For example, four effective search operators (i.e., accuracy-preferred domination, quick bit mutation operator, mutation-retry, and combination operators) designed by Zhou *et al.* [6] under the Pareto-based domination framework [38] can accelerate the convergence speed and strengthen the search capacity of FS. A steering matrix based on symmetric uncertainty and the domination relationship metric proposed by Cheng *et al.* [7] helps to select relevant features according to the importance of each feature. A duplication analysis-based EA reported by Xu *et al.* [39] applies a duplication method to filter out redundant features and a diversity-based selection method to guide the selection of promising features.

In particular, PSO has become a very popular solver for handling high-dimensional FS problems due to its efficient search capability and easy implementation. For example, a new potential PSO algorithm (PPSO) proposed by Tran *et al.* [15] uses a new discretized representation strategy to narrow the search space for high-dimensional FS problems. To further enhance the performance of PPSO, an effective variable-length representation strategy designed by Tran *et al.* [16] enables different particles in a swarm to have different variable lengths (i.e., the variable length of a particle equals the number of features it contains). Thus, the strategy can reduce the search space and help to explore different potential regions. In addition, to combine the distinct advantages of different FS methods, some PSO-based hybrid strategies were designed. For instance, an efficient hybrid PSO with a wrapped approach proposed by Chen *et al.* [40] applies an adaptive PSO variant to accelerate the convergence speed and a spiral-shaped mechanism to explore the search space and find promising regions. An effective filter-clustering-evolutionary wrapper algorithm (HFS-C-P) proposed by Song *et al.* [12] employs a clustering method to significantly reduce the search space and an evolutionary wrapper method with strong global search capability to find more representative features. However, for solving certain complex high-dimensional FS problems, the above PSO-based FS methods have some limitations, such as slow convergence speed and insufficient search capability in large-scale search spaces [13], [21].

To address the above issues, the idea of extending EMT for solving high-dimensional FS problems has been proposed, given by the superiority of EMT in solving certain complex optimization problems [41]-[43]. Two representative works were proposed by Chen *et al.* in [13], [21]. Specifically, Chen *et al.* first implemented the PSO-EMT, an EMT technique, to handle high-dimensional FS problems in [13]. In this approach, all features are classified into important and unimportant features by using a single filtering method (i.e., the *Relief-F* method [3]). In this way, one FS task with the relevant features and one FS task with all features are constructed. Then, these two relevant FS tasks are solved simultaneously by using a modified PSO variant that relies on knowledge transfer. Furthermore, Chen *et al.* designed an effective multiple task generation strategy in MTPSO [21], which not only considers the relevant features but also gives a selection probability for partially irrelevant features to form relevant FS tasks, considering the interactions between features.

Although the empirical results in [13], [21] have validated the effectiveness of employing EMT to solve high-dimensional FS problems, as introduced in **Section I**, the multiple task generation methods in PSO-EMT and MTPSO are rather simple, as they only use the *Relief-F* method to collect the relevant features, which cannot provide more diversified tasks for knowledge transfer and may be not applicable to different datasets with different preferences. In addition, the multiple task optimization strategies used in PSO-EMT and MTPSO are both based on PSO, which has limitations in terms of con-vergence speed and search ability when solving certain complex high-dimensional FS problems. Based on the above discussions, this paper proposes a more effective EMT-based FS method (MF-CSO) that improves the performance of both multiple task generation and optimization. Thus, this paper designs a diversified multiple task generation strategy with multiple filtering methods to construct several diversified relevant tasks, which is different from the existing EMT-based FS methods (i.e., PSO-EMT and MTPSO) using only a single filtering method. In this way, our method can transfer useful search knowledge among different relevant tasks, which enhances the ability to handle high-dimensional FS problems. Moreover, this paper modifies a CSO variant to run knowledge transfer among relevant tasks, aiming to overcome the shortcomings of PSO mentioned above. The details of the proposed EMT-based FS method are introduced in the next section.

## III. PROPOSED METHOD

This section introduces the details of our EMT-based FS method (MF-CSO). Specifically, **Section III. A** describes the general framework of MF-CSO, **Section III. B** introduces the process of multiple task generation in MF-CSO, **Section III. C** presents the principle of the multiple task optimization strategy that uses the proposed CSO-based knowledge transfer to evolve several relevant tasks, **Section III. D** describes the adopted fitness function in MF-CSO, and **Section III. E** analyzes the computational complexity of MF-CSO.

### A. General Framework of MF-CSO

As in most existing FS algorithms [6]-[7], the given dataset is first divided into a training set and a testing set. Then, a set of optimal solutions is obtained based on the training set using our proposed MF-CSO. Finally, the optimal solution is loaded into the testing set to evaluate the performance of the algorithm. The schematic of the proposed MF-CSO is plotted in Fig. 1, which includes three main steps (multiple task generation, multiple task optimization and result output). Specifically, the process of multiple task generation is first implemented by using $K$ filtering methods. Thus, $K+1$ highly relevant FS tasks are constructed, consisting of one FS task with all features and $K$ FS tasks with some important features selected by using $K$ filtering methods. The details of the proposed multiple task generation strategy are introduced in **Section III. B**. Then, the multiple task optimization process is conducted to solve these relevant tasks simultaneously, in which a modified CSO variant is used to perform knowledge transfer among the relevant tasks. The details of our proposed CSO-based knowledge transfer strategy for multiple task optimization are provided in **Section III. C**. In the last step, the particle with the best fitness function value for each task is output when the termination condition is met. The adopted fitness function is introduced in **Section III. D**. The pseudocode of the general framework of MF-CSO is provided in **Algorithm 1.** After the above training process, the final features selected in the best performing particle among $K+1$ tasks are used to assess the classification accuracy of the proposed MF-CSO method on the testing dataset.

### B. Multiple Task Generation Strategy with Multiple Filtering Methods

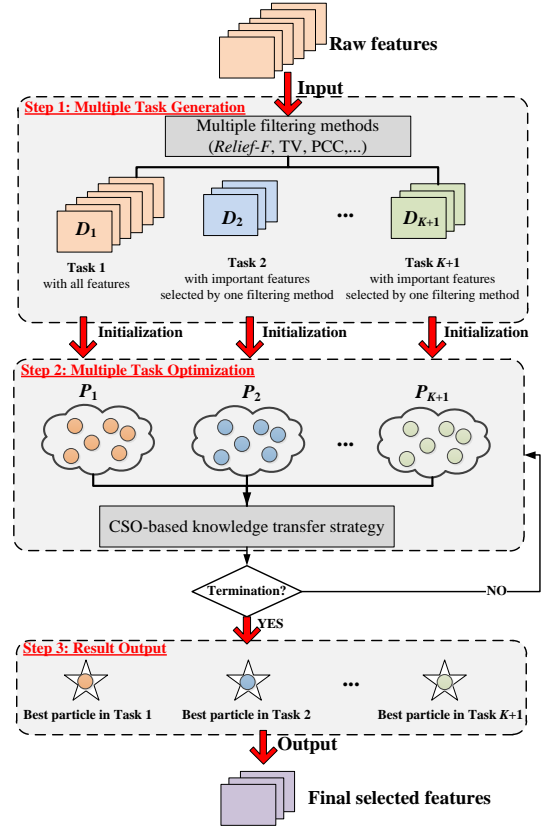

Fig. 1. Schematic of MF-CSO.

**Algorithm 1** Framework of MF-CSO

**Input:** *Data*: training set.
**Output:** *B*: the best particles of all tasks.
1. $(D_1, D_2,...,D_{K+1}) \leftarrow$ **Multiple Task Generation** (*Data*);
2. Initialize $K+1$ populations $(P_1, P_2,...,P_{K+1})$ for all tasks;
3. **while** the termination condition is not met **do**
4.    **Multiple Task Optimization** ($P_i$); \\ $i=1,2,...,K+1$
5. **end while**
6. $B \leftarrow$ Select the best particle from all tasks;

As mentioned in many existing FS studies [12]-[13], the filtering method is an effective technique to measure the importance of features, which helps to achieve good classification performance by eliminating redundant/irrelevant features. In general, different filtering methods show distinct advantages on different datasets due to their different metrics for calculating the correlation between features and labels. Thus, this paper uses multiple filtering methods to design a diversified multiple task generation strategy, aiming to enhance the robustness of our approach in handling various high-dimensional datasets.

First, the importance of features is ranked by using multiple filtering methods. Notably, our designed framework is scalable for most existing filtering methods. In particular, three common filtering methods are adopted in this paper to determine the importance of features, namely, the *Relief-F* [3], term variance (TV) [5], and Pearson correlation coefficient (PCC) methods [4]. The details of each filtering method are introduced as follows:

In the *Relief-F* method, the feature weight values are calculated based on the ability of the features to identify nearby samples. Specifically, *Relief-F* randomly selects a sample $R_i$ from the training set and then searches for $h$ nearest neighbors of $R_i$ from the same class, called the nearest hits $H_j$ ($j=1,2,...,h$). All the $h$ nearest neighbors from each of the different classes are called the nearest misses $M_j(c)$, where $c$ is a class. Then, the weight values of features based on the *Relief-F* method are calculated by:

$$W_{Relief-F}(a)^{t+1} = W_{Relief-F}(a)^t - \frac{\sum_{j=1}^{h} diff(a,R_i,H_j)}{t \times h} + \frac{\sum_{c \notin class(R_i)}[\frac{p(c)}{1-p(class(R_i))} \times \sum_{j=1}^{h} diff(a,R_i,M_j(c))]}{t \times h} \quad (5)$$

where $W_{Relief-F}(a)^t$ is the weight value of feature $a$ using the *Relief-F* method, $t$ is the current iteration, $h$ represents the size of the nearest neighbor samples, $class(R_i)$ indicates the class label for sample $R_i$, and $p(c)$ and $p(class(R_i))$ are the proportions of class $c$ and sample $R_i$, respectively. $diff(a,S_1,S_2)$ represents the difference between the values of feature $a$ in sample $S_1$ (i.e., $S_1(a)$) and sample $S_2$ (i.e., $S_2(a)$), calculated by:

$$diff(a,S_1,S_2) = |S_1(a) - S_2(a)|/(\max(a) - \min(a)) \quad (6)$$

where $\max(a)$ and $\min(a)$ indicate the maximum and minimum values of feature $a$ in the $h$ nearest samples of $H_j$ or $M_j(c)$, respectively. Note that higher weight values correspond to a higher relevance of these features.

The TV method is considered one of the simplest ways to assess the importance of features on a univariate basis as follows:

$$W_{TV}(a) = \frac{1}{|S|} \sum_{j=1}^{|S|} (A(j) - \overline{A})^2 \quad (7)$$

where $W_{TV}(a)$ is the weight value of feature $a$ based on the TV method, $A(j)$ represents the value of feature $a$ in sample $j$, and $\overline{A}$ represents the average value of all $A(j)$ ($j = 1, 2,..., |S|$, and $|S|$ indicates the total number of samples). Thus, a larger value of $W_{TV}(a)$ indicates that feature $a$ contains valuable information, which is more important for classification problems.

The PCC method ranks the importance of features by calculating the linear correlations between individual features and class labels in classification or continuous targets in regression, defined by

$$W_{PCC}(a) = \frac{\sum_{i=1}^{|S|}(x_i - \overline{x})(y_i - \overline{y})}{\sqrt{\sum_{i=1}^{|S|}(x_i - \overline{x})^2 \sum_{i=1}^{|S|}(y_i - \overline{y})^2}} \quad (8)$$

where $W_{PCC}(a)$ is the weight value of feature $a$ calculated by the PCC method. $x_i$ and $y_i$ represent the value of feature $a$ and its corresponding label in sample $i$, respectively. $\overline{x}$ and $\overline{y}$ represent the average values of $x_i$ and $y_i$, respectively. $|S|$ is the size of the samples. Note that $W_{PCC}(a) = 0$ means that the two variables (i.e., $x_i$ and $y_i$) are uncorrelated, $W_{PCC}(a) < 0$ means that the two variables are negatively correlated, and $W_{PCC}(a) > 0$ indicates that the two variables are positively correlated.

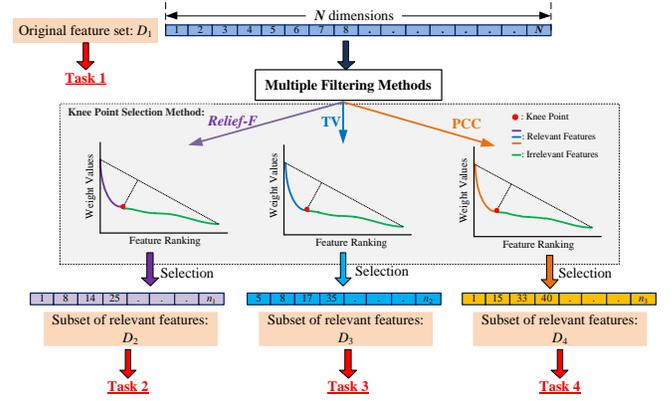

Fig. 2. An example of the proposed multiple task generation using multiple filtering methods.

Therefore, a larger value of $W_{PCC}(a)$ implies that feature $a$ is more correlated with the label.

After determining the importance of each feature based on the aforementioned multiple filtering methods, this paper employs the knee point selection method [44] based on the weight values of features to distinguish relevant and irrelevant features. That is, if and only if the weight values of features are larger than that of the knee point are these features considered relevant features, and vice versa. Due to page limitations, additional details of the adopted knee-point selection method are presented in the supplementary file.

For a more intuitive view of the entire multiple task generation process, a simple example is plotted in Fig. 2, where three different subsets of relevant features determined by three filtering methods (the *Relief-F*, TV, and PCC methods) are obtained, namely, $D_2$, $D_3$ and $D_4$, respectively. Finally, these feature subsets are used to construct three low-dimensional FS tasks, namely, Task 2, Task 3 and Task 4. In addition, a full set of features is used to construct an FS task, denoted as Task 1. In this way, we can obtain several different FS tasks through our proposed multiple task generation strategy using multiple filtering methods, helping to enhance the performance of knowledge transfer in EMT when handling various kinds of high-dimensional datasets.

*C. Multiple Task Optimization using a Competitive Swarm Optimizer*

After the multiple task generation process is completed, the multiple task optimization process is performed on these four relevant FS tasks, including Task 1 with all features and Tasks 2-4 with the important features determined by the *Relief-F*, TV and PCC methods, respectively. At the beginning of multiple task optimization, four subpopulations, i.e., $P_1, P_2, P_3$ and $P_4$, are initialized to solve their corresponding tasks simultaneously.

In terms of the solver in EMT for solving FS problems in high-dimensional classification, *Chen et al.* [13], [21] extended PSO to perform knowledge transfer between different tasks. However, as analyzed in **Section I**, PSO-based solvers still suffer from slow convergence speeds or even premature convergence when handling complex FS problems in large-scale search spaces. In general, there are two main approaches to efficiently solve high-dimensional problems in the EC domain. Specifically, the first approach simplifies the problems to be solved by using divide-and-conquer approaches [45]-[48]. Nevertheless, as described in [49], such approaches are computationally expensive because they consume a large amount of additional computational resources for variable analysis or dimensionality reduction. In addition, as pointed out in [50], such approaches are more likely to become trapped in local optima since they fail to fully explore the original large-scale space. The second approach designs more powerful search operators to effectively explore large-scale search spaces. CSOs are a representative of this approach and show strong exploration ability, and the empirical results in [36]-[37] demonstrate their superior advantages. Thus, a natural idea is using CSO within EMT to improve the exploration ability for solving high-dimensional FS problems. This paper introduces a new CSO variant as the solver for EMT.

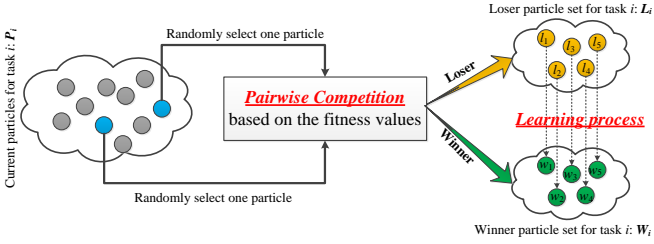

Fig. 3. Schematic of pairwise competition and learning process for competitive swarm optimizer.

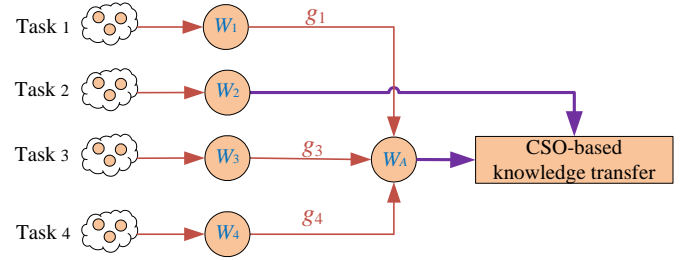

Fig. 4. A simple example of the CSO-based knowledge transfer method for optimizing Task 2, where $g_1 + g_3 + g_4 = 1$.

**Algorithm 2** presents the pseudocode for our designed multiple task optimization method using CSO. First, as shown in Line 2, the particles in each task are divided into winner particles and loser particles by using the random pairwise competition of CSO. Based on the principle of pairwise competition, two particles are randomly selected from the swarm and then compared based on the fitness values. Then, the particle with the better fitness value is inserted into the winner particle set $W_i$, while the other particle is inserted into the loser particle set $L_i$. the calculation of the fitness value is introduced in **Section III. D**. To better understand the pairwise competition in CSO, a schematic diagram is provided in Fig. 3. Then, the CSO-based knowledge transfer strategy is performed to evolve each task in Lines 3-10, consisting of two cases (i.e., knowledge transfer or not). Specifically, a real-valued number randomly generated in (0, 1) is compared to the predefined knowledge transfer probability $p_{trans}$. If this random number is less than $p_{trans}$, knowledge transfer will not occur, and the traditional CSO search will be performed to evolve the loser particles, where the velocity and position of loser particles are updated by learning from their corresponding winner particles using (4)-(5). Otherwise, knowledge transfer will be realized by using the modified CSO, where the loser particle learns from not only its winner particle but also another winner particle obtained in other relevant tasks. The formulation is defined as follows:

$$V_L^i(t+1) = r_1 \times V_L^i(t) + r_2 \times (X_w^i(t) - X_L^i(t)) + r_3 \times (X_{W_A}(t) - X_L^i(t)) \quad (9)$$

where $t$ and $t+1$ are the current $t$-th generation and the next generation, respectively, and $r_1$, $r_2$ and $r_3$ are three random numbers ranging from 0 to 1. $X_w^i(t)$ and $X_L^i(t)$ represent the current positions of winner and loser particles for task $i$, respectively, and $X_{W_A}(t)$ represents the position of one promising winner particle that is aggregated by three winner particles from the other three relevant tasks. The formulation is defined as follows:

$$X_{W_A}(t) = \sum_{k \in [1,4] \neq i} X_{W_k}(t) \times g_k \quad (10)$$

where $X_{W_k}(t)$ indicates the position of the winner particle that is randomly selected from task $k$ excluding the target task $i$. Fig. 4 gives a simple example of using CSO-based knowledge transfer for optimizing Task 2, where $g_k$ is the corresponding weight value of task $k$. In this paper, we set the weight value of Task 1, which contains the full set of features, to 0.1, and the weight value of other tasks containing different feature subsets to 0.45. The reasons for these parameter settings are discussed in **Section IV. D**.

After all the loser particles in each task have evolved as described, polynomial mutation (PM) [38] is used to evolve the winner particles in Line 11, aiming to further improve their quality. Then, all the updated winner and loser particles are combined to form the next subpopulation for each task in Line 12. Note that the multitasking optimization process will be terminated when the stopping condition is met.

*D. Fitness Function*

The design of fitness function values for evaluating the performance of each individual plays a crucial role in the field of FS, as a reasonable fitness function definition can effectively guide the FS process and select high-quality features. Essentially, the main objectives of FS are to maximize the classification accuracy and minimize the size of the feature subset. Therefore, in this paper, a fitness function consisting of these two objectives is designed as follows:

$$fitness = \alpha \times E + (1-\alpha) \times \frac{S}{D} \quad (11)$$

where $E$ is the classification error rate of the algorithm and $S$ and $D$ indicate the number of selected features and the total number of features, respectively. As suggested in [51], $\alpha$, ranging from 0 to 1, is used to dynamically adjust the weight value between the classification accuracy and the size of selected features, and we set $\alpha = 0.999999$ as advised in a previous study [11].

As the dimensionality of the dataset increases, the impact of the data imbalance problem becomes increasingly significant. Therefore, to better tackle the data imbalance problem, we use a balanced classification error rate instead of the general error rate, as recommended in [52]. The balanced classification error rate is calculated as follows:

$$Balanced\_E = 1 - \frac{1}{c}\sum_{i=1}^{c} TPR_i \quad (12)$$

where $Balanced\_E$ is the balanced error rate, $c$ denotes the number of classes in the dataset and $TPR_i$ is the proportion of correctly identified instances in class $i$. Since the balanced accuracy has no bias for each class in FS problems, the weight for all classes is set to $1/c$.

*E. Computational Complexity Analysis*

As presented in **Algorithm 1**, MF-CSO consists of four main components: the process of multiple task generation, the process of multiple task optimization and the result output. Specifically, the process of multiple task generation is run only at the beginning of the algorithm with the input of the training set with $D$ features, which requires a time complexity of $O(K \times D)$, where $K$ is the number of adopted filtering methods in MF-CSO. Then, the process of population initialization is performed in Line 2, which requires a total time complexity of $O(K \times N)$, where $N$ is the total population size. The process of multiple task generation with the input of $K+1$ different tasks has a time complexity of $(K+1) \times O(D \times N_i)$, where $N_i$ is the subpopulation size for task $i$. The process of result output has a time complexity of $(K+1) \times O(D \times N_i)$. Therefore, the worst time complexity of our proposed MF-CSO is $Max\{O(D \times K), O(K \times N), O(D \times N_i)\}$.

IV. SIMULATION RESULTS AND DISCUSSION

This section presents a large number of experiments to empirically validate the performance of our proposed MF-CSO in processing eighteen real-world high-dimensional datasets with the number of features ranging from 2000 to 12000. In addition, ablation experiments further demonstrate the effectiveness of our designed strategies in MF-CSO. All the simulations are performed by using MATLAB R2020a on a 64-bit Windows 10 personal computer with 24 GB RAM and an Intel Core-i7 3.6 GHz processor.

*A. Comparison Algorithms*

To verify the effectiveness of the proposed MF-CSO, four competitive EA-based FS methods, including PS-NSGA [6], SM-MOEA [7], PSO-EMT [13], and MTPSO [21], are adopted for performance comparison. Specifically, SM-MOEA and PS-NSGA are wrapper-based FS methods, while PSO-EMT and MTPSO use the idea of EMT for solving high-dimensional FS problems, which have shown promising performance in high-dimensional classification. In addition, two traditional EA-based FS methods, MOEA/D-FS [53] and CSO-FS [22], are used for performance comparison. Particularly, MOEA/D-FS uses a binary crossover variation in the framework of MOEA/D [53], and CSO-FS employs a traditional CSO search strategy [22] for handling FS problems. Moreover, to verify the effectiveness of these FS methods, the classifica-tion effect without the FS method is adopted for performance comparison, termed FULL. Furthermore, three traditional filtering methods (e.g., the *Relief-F* [3], PCC [4] and TV [5] methods) are also used for performance comparison. Note that for a fair comparison, the knee point selection strategy [44] is also used for these three filtering methods.

TABLE I: PROPERTIES OF THE ADOPTED DATASETS.

| No. | Dataset | Features | Instances | Classes |
|---|---|---|---|---|
| 1 | warpPIE10P | 2,420 | 210 | 10 |
| 2 | Lymphoma | 5,026 | 62 | 3 |
| 3 | Nci | 5,244 | 61 | 8 |
| 4 | Leukemia 1 | 5,327 | 72 | 3 |
| 5 | DLBCL | 5,469 | 77 | 2 |
| 6 | 9Tumor | 5,726 | 60 | 9 |
| 7 | TOX_171 | 5,748 | 171 | 4 |
| 8 | Brain Tumor 1 | 5,920 | 90 | 5 |
| 9 | Prostate6033 | 6,033 | 102 | 2 |
| 10 | ALLAML | 7,129 | 72 | 2 |
| 11 | Nci9 | 9,712 | 60 | 9 |
| 12 | Adenocarcinoma | 9,868 | 76 | 2 |
| 13 | orlraws10P | 10,304 | 100 | 10 |
| 14 | Brain Tumor 2 | 10,367 | 50 | 4 |
| 15 | Prostate | 10,509 | 102 | 2 |
| 16 | Leukemia 2 | 11,225 | 72 | 3 |
| 17 | 11 Tumor | 12,533 | 174 | 11 |
| 18 | Lung Cancer | 12,600 | 203 | 5 |

TABLE II: PARAMETER SETTINGS

| Algorithms | Parameters |
|---|---|
| SM-MOEA | Attenuation factor $\gamma$=0.1 |
| PS-NSGA | Mutation probability=0.1, Mutation retry number=1 |
| PSO-EMT | $c_1$=$c_2$=$c_3$=1.49445, $\rho$=0.05, $rmp$=0.6, $m$=10, $w = 0.9 - 0.5 \times (iter/max\_iter)$ |
| MTPSO | $c_1$=$c_2$=$c_3$=1.49445, $\rho$=0.05, $rmp$=0.6, $G$=6, $w = 0.9 - 0.5 \times (iter/max\_iter)$ |
| CSO-FS | $r_1$, $r_2$, $r_3 \in [0,1]$ |
| MOEA/D-FS | $T$=0.8, $\rho$=0.5 |
| MF-CSO | $r_1$, $r_2$, $r_3 \in [0,1]$, $g_1$=0.1, $g_2$=$g_3$=$g_4$=0.45, $p_{trans}$=0.5 |

All the source codes of the compared algorithms, except PSO-EMT and MTPSO, are provided by the original references. To verify the experimental results and to facilitate further study by other researchers, we publish the source codes of our method and two reproduction methods (i.e., PSO-EMT and MTPSO) on GitHub[1].

*B. Datasets*

In our experiments, eighteen unbalanced high-dimensional genetic datasets with more than 2000 features are adopted to validate the effectiveness of the proposed MF-CSO, which are also available in[1]. Table I summarizes the properties of these adopted datasets.

---
[1] https://github.com/gelin123/MF_CSO_Materials

*C. Parameter Settings*

To ensure fairness, a KNN classifier with *k*=1 is chosen for all the compared FS methods. As suggested in [6], the maximum number of iterations and the population size are set to 70 and 300, respectively, for all the compared methods. Notably, the three hybrid methods, i.e., PSO-EMT, MTPSO and MF-CSO, use a discrete coding approach. Therefore, during the evolutionary search process, a classification rule based on a threshold $\delta$ determines the selected features. Specifically, if the particle representation corresponding to a feature is greater than $\delta$, the feature is marked as '1' (i.e., the feature is selected); otherwise, it is marked as '0' (i.e., the feature is not selected). In this paper, $\delta$ is set to 0.5 for these three discrete coding-based algorithms. In addition, Table II summarizes some unique parameters of each compared algorithm.

To ensure the fairness and validity of the comparison results, the tenfold cross-validation (CV) method is used to estimate the results of most existing FS algorithms [6]-[7]. In addition, to avoid feature selection bias, a tenfold CV method is performed on the training sets during the evolutionary process, which aims to group the training sets to evaluate the objective value of the current solution.

*D. Parameter Sensitivity Analysis of MF-CSO*

As introduced in **Algorithm 2**, our designed MF-CSO consists of two important parameters, including the transfer probability $p_{trans}$ that determines the evolutionary method (i.e., traditional CSO or knowledge transfer-based CSO) and the weight value $g_k$ that controls the impact of each task when conducting knowledge transfer among tasks. To clarify the reasons for the chosen parameter settings, we conduct a parameter sensitivity analysis of these two parameters.

1). *Sensitivity Analysis of $p_{trans}$:* The transfer probability $p_{trans}$ plays a critical role in evolutionary multitasking optimization. Here, we perform a parameter sensitivity analysis to determine the appropriate parameter setting for $p_{trans}$. Fig. 5 provides the error rates of MF-CSO with different values of $p_{trans}$ on the 'warpPIE10P' (with a small number of features), 'Leukemia 1' (with a medium number of features) and 'Leukemia 2' (with a large number of features) datasets, where the best result on each dataset is marked by a red pentagram (★). As observed from Fig. 5, MF-CSO with various values of $p_{trans}$ indeed shows different performance when processing different datasets. In summary, MF-CSO with $p_{trans}$=(0.4, 0.5, 0.6) shows a better overall performance compared to the others. Therefore, the transfer probability $p_{trans}$ is suggested to be set between 0.4 and 0.6, and $p_{trans}$=0.5 is adopted in this study.

2). *Sensitivity Analysis of $g_k$:* When setting the weight values for the four different tasks in MF-CSO (i.e., $g_1$ for Task 1 with a full set of features and $g_2$, $g_3$, and $g_4$ for Tasks 2-4 with different subsets of features), some principles are considered. First, the weight values of Tasks 2-4 should be equal (i.e., $g_2 = g_3 = g_4$) to ensure that Tasks 2-4 have the same influence on the knowledge transfer process when optimizing Task 1. Second, the weight value of Task 1 should be smaller than that of Tasks 2-4 (i.e., $g_1 < g_{2,3,4}$). The reason behind this rule is that it is easier to find high-quality solutions in a small-scale search space than in the original large-scale search space. Therefore, the particles of Tasks 2-4 should be more influential than those of Task 1 when performing knowledge transfer. Third, based on the aggregation mechanism designed in (10), $g_2$, $g_3$, and $g_4$ work when optimizing Task 1, $g_1$, $g_3$, and $g_4$ work when optimizing Task 2, and so on. Therefore, the sum of the weight values between Task 1 and any other two tasks are equal to 1 (i.e., $g_1 + g_i + g_j = 1, i, j \in \{2,3,4\} \wedge i \neq j$).

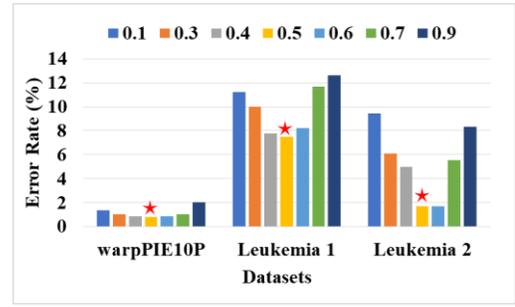

Fig. 5：The average error rates of MF-CSO with different transfer probabilities on the 'warpPIE10P', 'Leukemia 1', and 'Leukemia 2' datasets.

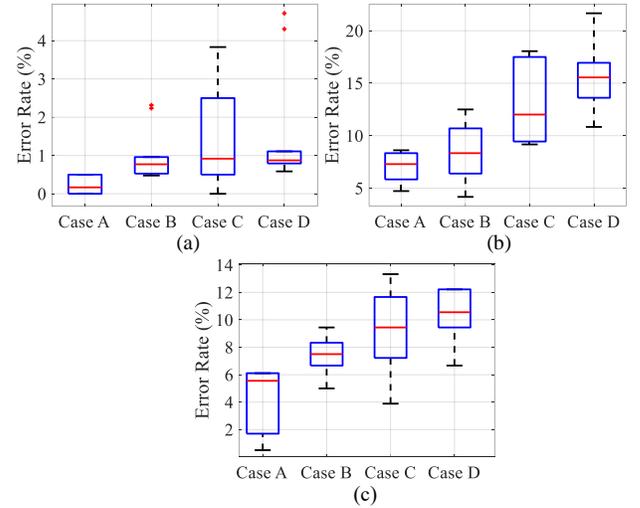

Fig. 6：The average error rates of MF-CSO with different weight value settings for the (a) 'warpPIE10P', (b) 'Leukemia 1', and (c) 'Leukemia 2' datasets.

Based on the above principles, four different sets for setting $g_1$ and $g_{2,3,4}$ are considered here for performing parameter sensitivity analysis, namely, case **A** ($g_1 = 0.1$, $g_2 = g_3 = g_4 = 0.45$), case **B** ($g_1 = 0.2$, $g_2 = g_3 = g_4 = 0.4$), case **C** ($g_1 = 0.33$, $g_2 = g_3 = g_4 = 0.33$), and case **D** ($g_1 = 0.6$, $g_2 = g_3 = g_4 = 0.2$). Fig. 6 provides the average error rates of these four cases in solving the 'warpPIE10P', 'Leukemia 1' and 'Leukemia 2' datasets. As observed from Fig. 6, compared to other cases, case **A** achieves the lowest error rate on all three datasets. Therefore, we adopt $g_1 = 0.1$ and $g_2 = g_3 = g_4 = 0.45$ in this study.

*E. Performance Comparison and Discussion*

TABLE III: THE AVERAGE CLASSIFICATION RESULTS OF MF-CSO VERSUS SEVEN COMPARED ALGORITHMS.

| Datasets | FULL | SM-MOEA | PS-NSGA | PSO-EMT | MTPSO | CSO-FS | MOEA/D-FS | MFCSO |
|---|---|---|---|---|---|---|---|---|
| warpPIE10P | 15.49(-) | 1.48(≈) | 1.43(≈) | 0.83(≈) | 0.84(≈) | 51.33(-) | 1.00(≈) | **0.82** |
| Lymphoma | **0.92**(+) | 23.59(-) | 7.99(-) | 6.67(-) | 3.33(-) | 43.62(-) | 2.20(≈) | 1.87 |
| Nci | 31.74(-) | 34.51(-) | 32.83(-) | 40.32(-) | 36.01(-) | 35.71(-) | 30.42(≈) | **28.77** |
| Leukemia 1 | 20.28(-) | 22.12(-) | 12.89(-) | 13.89(-) | 13.06(-) | 17.50(-) | 17.31(-) | **7.50** |
| DLBCL | 17.00(-) | 19.63(-) | 15.95(-) | 15.83(-) | 11.67(-) | 15.50(-) | 15.83(-) | **7.50** |
| 9Tumor | 63.33(-) | 52.64(+) | 45.85(+) | **44.41**(+) | 47.43(+) | 56.06(≈) | 56.17(≈) | 55.23 |
| Tox_171 | 22.11(-) | 16.11(-) | 8.33(≈) | 8.00(≈) | 8.50(≈) | 14.21(-) | 10.63(-) | **7.62** |
| Brain Tumor 1 | 27.92(-) | 25.65(≈) | 28.76(-) | 22.33(≈) | **21.97**(+) | 26.70(-) | 24.84(≈) | 24.33 |
| Prostate6033 | 18.69(-) | 18.28(-) | 16.03(-) | 19.50(-) | 16.00(-) | 15.58(-) | 19.67(-) | **12.82** |
| ALLAML | 22.54(-) | 18.83(-) | 13.11(-) | 9.10(-) | 8.92(-) | 17.22(-) | 17.00(-) | **3.76** |
| Nci9 | 58.67(-) | **38.42**(+) | 43.90(-) | 46.95(≈) | 48.18(-) | 56.81(-) | 54.58(-) | 46.00 |
| Adenocarcinoma | 37.26(-) | 30.40(≈) | 36.06(≈) | 34.67(≈) | 33.98(≈) | 35.99(≈) | 40.48(-) | 34.44 |
| orlraws10P | 22.18(-) | 6.28(-) | 6.70(-) | 3.40(≈) | **3.00**(≈) | 7.67(-) | 5.00(≈) | 3.70 |
| Brain Tumor 2 | 37.50(-) | 42.79(-) | 33.67(-) | 33.33(-) | 26.67(-) | 27.60(≈) | 36.00(-) | **25.42** |
| Prostate | 14.67(≈) | 17.19(-) | 12.15(≈) | 18.24(-) | 14.83(≈) | **11.03**(≈) | 16.72(-) | 12.67 |
| Leukemia2 | 10.56(-) | 16.86(-) | 10.94(-) | 9.44(-) | 10.00(-) | 14.15(-) | 12.40(-) | **1.67** |
| 11 Tumors | 28.58(-) | 29.71(-) | 22.19(≈) | 21.46(≈) | 20.00(≈) | 27.61(-) | 25.62(-) | **20.31** |
| Lung Cancer | 21.95(-) | 12.94(≈) | 13.50(≈) | 15.40(-) | 15.72(-) | 21.55(-) | 21.17(-) | **12.44** |

TABLE IV: THE NUMBERS OF FEATURES SELECTED BY MF-CSO AND SEVEN COMPARED ALGORITHMS.

| Datasets | FULL | SM-MOEA | PS-NSGA | PSO-EMT | MTPSO | CSO-FS | MOEA/D-FS | MFCSO |
|---|---|---|---|---|---|---|---|---|
| warpPIE10P | 2420 | 18 | 162.5 | 115.47 | 227.76 | **4.6** | 1070.2 | 103.93 |
| Lymphoma | 4026 | 71 | **2.1** | 12 | 11.99 | 2.61 | 1896.7 | 53.48 |
| Nci | 5244 | **15.06** | 125.23 | 264.74 | 1467.47 | 403.17 | 2495.8 | 273.46 |
| Leukemia 1 | 5327 | **9.16** | 16.8 | 242.5 | 910.54 | 415.93 | 2535.4 | 229.6 |
| DLBCL | 5469 | 12.8 | **11.2** | 154.4 | 1235.27 | 343.77 | 2629.6 | 96.5 |
| 9Tumor | 5726 | **14.53** | 192.2 | 309.3 | 574.293 | 339.15 | 2753.9 | 199.2 |
| Tox_171 | 5748 | **18.21** | 94.4 | 2867.8 | 879.863 | 585.43 | 2744.5 | 1170.7 |
| Brain Tumor 1 | 5920 | **6.4** | 61.11 | 211.5 | 984.59 | 139.28 | 2823 | 194.5 |
| Prostate6033 | 6033 | **9.89** | 46.52 | 342.336 | 1625.71 | 710.92 | 2857.3 | 212.13 |
| ALLAML | 7129 | **5.33** | 17.9 | 182.8 | 1955.12 | 401.87 | 3414.5 | 297.6 |
| Nci9 | 9712 | **26.3** | 167.7 | 1333.7 | 689.98 | 560.133 | 4669.1 | 2354.4 |
| Adenocarcinoma | 9868 | **11.49** | 55.95 | 2084.56 | 297.77 | 533.773 | 4660.5 | 523.31 |
| orlraws10P | 10304 | 28.9 | **27.93** | 807.88 | 1399.04 | 57.7 | 4899.4 | 322.51 |
| Brain Tumor 2 | 10367 | **9.27** | 74.66 | 211.5 | 1909.09 | 139.28 | 4987.8 | 194.5 |
| Prostate | 10509 | **13.5** | 63.2 | 132.1 | 2880.71 | 1203.74 | 5065.88 | 63.2 |
| Leukemia2 | 11225 | **8.56** | 28.06 | 244.3 | 1605.6 | 1109.66 | 5420.7 | 373.4 |
| 11 Tumors | 12533 | **46.44** | 334.72 | 884.3 | 2143.44 | 618.45 | 6096.3 | 334.72 |
| Lung Cancer | 12600 | **20.69** | 106.3 | 609.63 | 723.4 | 1163.09 | 6027.6 | 498.1 |

In this section, a large number of experiments are conducted to verify the effectiveness of MF-CSO on eighteen real-world high-dimensional classification datasets. Specifically, the experimental results in terms of the error rate and the number of selected features obtained by the proposed MF-CSO and several competitive EA-based FS methods are provided in Table III and Table IV, respectively, and the experimental results obtained by MF-CSO and three traditional filtering methods are summarized in Table V. In addition, the results of Wilcoxon significance tests with a significance level of 0.05 are presented in column *S*, where '+', '-' and '≈' indicate that the compared method is better than, worse than, and similar to the proposed MF-CSO in terms of the error rate, respectively. All the experimental results are obtained from 30 independent runs of tenfold cross-validation where the selected folds and the algorithm stochastic depend on the selected random seed, and the lowest error rate in each dataset is marked by **boldface**.

1) *Comparison with State-of-the-art EA-based FS Methods*

In terms of the error rates listed in Table III, MF-CSO achieves the best results in 11 out of 18 datasets. Particularly, MF-CSO shows obvious advantages over other competitors on five datasets, specifically, the 'Leukemia 1', 'DLBCL', 'Prostate6033', ' ALLAML and 'Leukemia 2' datasets, as the error rates of MF-CSO on these five datasets are much lower than those of the other algorithms. In particular, the superiority of MF-CSO over non-EMT-based methods, i.e., FULL, SM-MOEA, PS-NSGA, CSO-FS and MOEA/D-FS, validates the effectiveness of using an efficient EMT technique. On the other hand, compared to two EMT-based FS methods (i.e., PSO-EMT and MTPSO), MF-CSO shows clear advantages in most cases. Thus, the superior performance of MF-CSO over PSO-EMT and MTPSO verifies the effectiveness of the designed multiple task generation method and the modified CSO variant for solving high-dimensional FS problems. In addition, as summarized in the last column $S$, compared to the other seven competitive EA-based FS methods, our MF-CSO wins 84 times, loses 8 times and draws 34 times out of all 126 comparisons in terms of the test error rate.

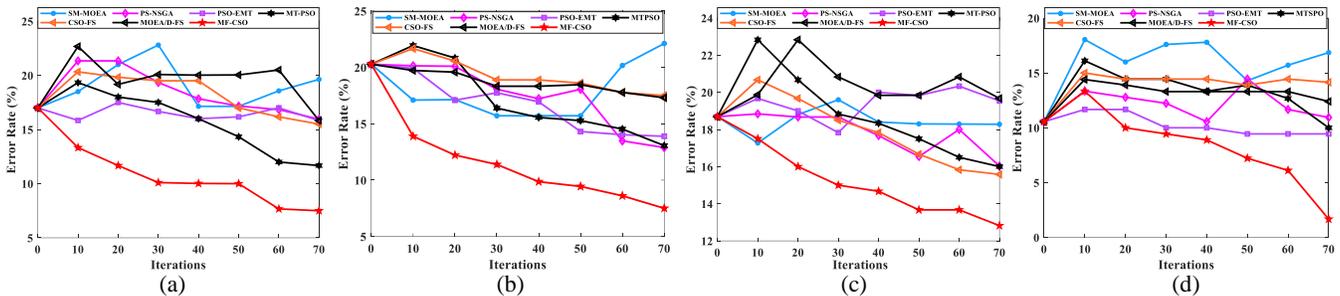

Fig. 7: The convergence profiles of compared algorithms over iterations on the (a) 'DLBCL', (b) 'Leukemia 1', (c) 'Prostate6033' and (d) 'Leukemia 2' datasets, respectively.

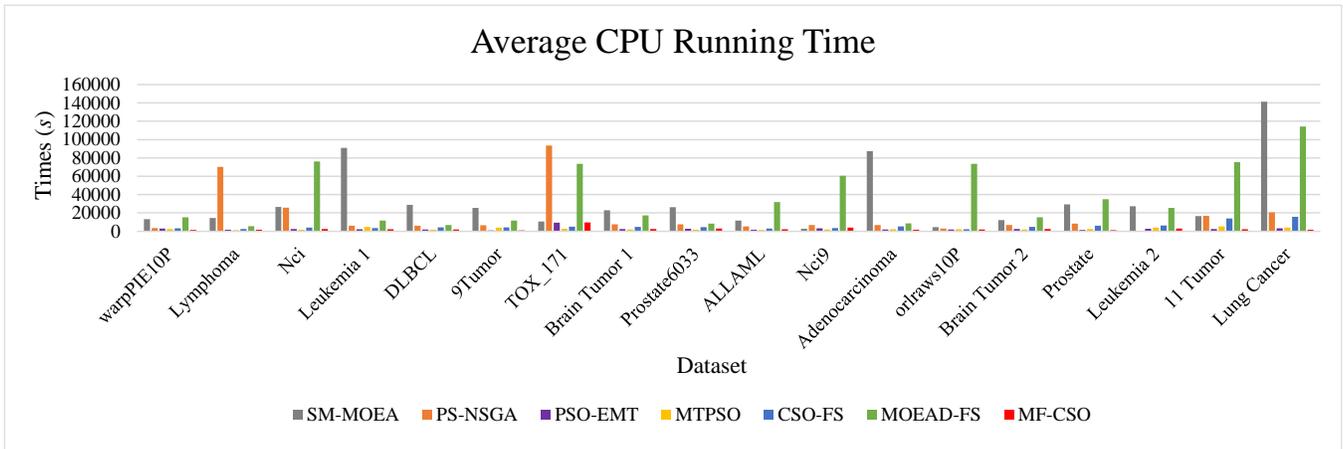

Fig. 8: The average CPU running time of the proposed MF-CSO and other EA-based FS methods over 30 independent runs.

Furthermore, considering the number of selected features in Table IV, the number of the selected features obtained by SM-MOEA and PS-NSGA are less than that by MF-CSO for most of the adopted datasets because the former design specialized strategies for fast dimensionality reduction. Specifically, SM-MOEA adopts a dimensionality reduction operator to recursively downsize the number of selected features, and PS-NSGA uses a variation operator to achieve fast dimensionality reduction. Although these two methods perform well in terms of dimensionality reduction, this feature will lead to a higher error rate in classification. As seen in Table IV, the error rates of MF-CSO are much lower than those of SM-MOEA and PS-NSGA in all adopted datasets except three datasets (i.e., the '9 Tumors', 'Nci 9' and 'Adenocarcinoma' datasets). When compared with other FS methods (i.e., FULL, PSO-EMT, MTPSO, CSO-FS and MOEA/D-FS), the numbers of selected features obtained by MF-CSO are the smallest on most adopted datasets.

In summary, the experimental results listed in Tables III-IV verify the effectiveness of the proposed MF-CSO compared to seven EA-based FS methods for solving these eighteen adopted datasets in terms of accuracy and number of selected features.

In addition, to visualize the convergence speed of each compared algorithm, the trends in error rates of each compared method over the iterations on the 'DLBCL', 'Leukemia 1', 'Prostate6033' and 'Leukemia 2' datasets are plotted in Figs. 7 (a)-(d), respectively. As learned from these figures, our proposed MF-CSO shows faster convergence speed and achieves better accuracy than the other compared methods.

*2) Comparison of CPU Running Time*

The training time is also an important criterion to evaluate the efficiency of an algorithm. Therefore, we further investigate the computational efficiency of MF-CSO and other EA-based FS methods. Fig. 8 provides the average CPU running time of MF-CSO and other EA-based FS methods over 30 independent runs. As shown in Fig. 8, MF-CSO shows the lowest running time in most of the datasets adopted. Specifically, MF-CSO shows similar runtimes to PSO-EMT, MTPSO, and CSO-FS, indicating that our multiple task generation and optimization strategies not only achieve good performance but also have low computational complexity. In addition, MF-CSO is significantly faster than the other three EA-based FS methods that do not use the EMT technique, namely, SM-MOEA, PS-NSGA, and MOEA/D-FS, which demonstrates the efficiency of using the EMT technique. The main reason behind this is that these three EA-based FS methods apply a multiobjective optimization framework for classifying high-dimensional datasets, which consumes a large amount of time to perform FS. In contrast, the EMT-based FS methods can generate promising FS tasks with a small number of features, which significantly reduces the computation time for each fitness evaluation during the FS process.

TABLE V: Comparison Between MF-CSO And Three Traditional Filtering Methods

| Dataset | Algorithm | Size | Error | S | Dataset | Algorithm | Size | Error | S |
|---|---|---|---|---|---|---|---|---|---|
| warpPIE10P | *Relief-F* | 401 | 0.50 | ≈ | ALLAML | *Relief-F* | 508 | 4.17 | ≈ |
|  | PCC | 282 | 1.5 | ≈ |  | PCC | 832 | 5.17 | ≈ |
|  | TV | 2242 | **0** | ≈ |  | TV | 660 | 28.67 | - |
|  | MF-CSO | 103.93 | 0.82 |  |  | MF-CSO | 297.6 | **3.76** |  |
| Lymphoma | *Relief-F* | 343 | 0.92 | ≈ | Nci9 | *Relief-F* | 9709 | 50.00 | - |
|  | PCC | 248 | 1.83 | ≈ |  | PCC | 9700 | 53.33 | - |
|  | TV | 521 | **0.84** | ≈ |  | TV | 7593 | 65.00 | - |
|  | MF-CSO | 53.48 | 1.87 |  |  | MF-CSO | 2354.4 | **46.00** |  |
| Nci | *Relief-F* | 548 | **16.43** | + | Adenocarcinoma | *Relief-F* | 3789 | 37.5 | - |
|  | PCC | 590 | 29.76 | ≈ |  | PCC | 9670 | 36.88 | - |
|  | TV | 397 | 24.77 | + |  | TV | 657 | 40.12 | - |
|  | MF-CSO | 273.46 | 28.77 |  |  | MF-CSO | 523.31 | **34.44** |  |
| Leukemia 1 | *Relief-F* | 436 | **6.33** | + | orlraws10P | *Relief-F* | 1013 | **0** | + |
|  | PCC | 784 | 10.44 | - |  | PCC | 1717 | 9.00 | - |
|  | TV | 421 | 11.81 | - |  | TV | 1161 | 8.00 | - |
|  | MF-CSO | 229.6 | 7.50 |  |  | MF-CSO | 322.51 | 3.70 |  |
| DLBCL | *Relief-F* | 347 | 9.17 | - | Brain Tumor 2 | *Relief-F* | 815 | 28.34 | - |
|  | PCC | 599 | 8.50 | ≈ |  | PCC | 916 | 38.34 | - |
|  | TV | 458 | 11.67 | - |  | TV | 989 | 27.50 | - |
|  | MF-CSO | 96.5 | **7.50** |  |  | MF-CSO | 194.5 | **25.42** |  |
| 9Tumor | *Relief-F* | 403 | **36.67** | + | Prostate | *Relief-F* | 686 | **10.00** | + |
|  | PCC | 899 | 56.67 | ≈ |  | PCC | 983 | 17.67 | - |
|  | TV | 380 | 45.00 | + |  | TV | 1242 | 19.50 | - |
|  | MF-CSO | 199.2 | 55.23 |  |  | MF-CSO | 63.2 | 12.67 |  |
| TOX_171 | *Relief-F* | 5474 | 14.75 | - | Leukemia2 | *Relief-F* | 1167 | 3.89 | - |
|  | PCC | 1371 | 26.59 | - |  | PCC | 1679 | 16.11 | - |
|  | TV | 319 | 27.00 | - |  | TV | 734 | 7.22 | - |
|  | MF-CSO | 1170.7 | **7.62** |  |  | MF-CSO | 373.4 | **1.67** |  |
| Brain Tumor 1 | *Relief-F* | 519 | 24.75 | ≈ | 11 Tumor | *Relief-F* | 829 | **17.87** | + |
|  | PCC | 837 | 21.25 | + |  | PCC | 1581 | 30.15 | - |
|  | TV | 611 | 29.17 | - |  | TV | 697 | 21.36 | ≈ |
|  | MF-CSO | 194.5 | **24.33** |  |  | MF-CSO | 334.72 | 20.31 |  |
| Prostate6033 | *Relief-F* | 518 | **10.00** | ≈ | Lung Cancer | *Relief-F* | 881 | 16.50 | - |
|  | PCC | 770 | 13.67 | ≈ |  | PCC | 2289 | 14.84 | - |
|  | TV | 657 | 17.83 | - |  | TV | 563 | 14.53 | - |
|  | MF-CSO | 212.13 | 12.82 |  |  | MF-CSO | 498.1 | **12.44** |  |

*3) Comparison with Traditional Filtering Methods*

Table V provides comparisons between the proposed MF-CSO and three traditional filtering methods, i.e., the *Relief-F*, PCC, and TV methods. Considering the error rate, MF-CSO obtains the best results in 9 out of 18 datasets. In particular, MF-CSO shows clear advantages on the 'Adenocarcinoma', 'Brain Tumor 2', 'Leukemia2' and 'Lung Cancer' datasets, as its error rates on these four datasets are much lower than those of the compared filtering methods. Moreover, when considering the performance on eight datasets with more than 9500 features (i.e., from the 'Nci 9' to the 'Lung Cancer' datasets), MF-CSO obtains the best results in 5 out of 8 datasets. Therefore, the superior performance of MF-CSO on such high-dimensional datasets indicates that our proposed EMT technique with multiple filtering methods is more advantageous for solving high-dimensional FS problems. In summary, among 54 comparisons in terms of the error rates, as shown in the last Column *S*, MF-CSO wins 30 times, loses 9 times and draws 15 times. In contrast, as listed in the left column of Table V, when considering the error rates on the datasets with fewer than 6000 features (i.e., from the 'warpPIE10P' to the 'Prostate6033' datasets), our proposed MF-CSO shows slight disadvantages or similar performance compared to three traditional filtering methods. The reason for this result is that transferred particles from other tasks may mislead the evolution of the task when solving these datasets with relatively few features, thus degrading the performance of MF-CSO. However, traditional filtering methods can perform an approximate screening of features by using a single metric to search for a promising feature subset and achieve good performance on datasets with a small number of features since the search space for these datasets is relatively small. It is worth noting that although the accuracies obtained by our proposed MF-CSO on these datasets with fewer than 6000 features are slightly worse than or similar to those of the *Relief-F*, PCC and TV methods, the number of features selected by MF-CSO is much smaller than that of these traditional filtering methods. In summary, the results in Table V demonstrate that the effectiveness of our proposed MF-CSO with multiple filtering methods becomes clearer as the number of features increases.

On the other hand, when considering the number of selected features, MF-CSO can achieve significant advantages compared to the *Relief-F*, PCC, and TV methods, MF-CSO selects the least number of features on all adopted datasets. Therefore, the proposed MF-CSO based on the idea of EMT shows advantages in terms of the error rate and the number of selected features on most datasets when compared to the three traditional filtering methods.

In summary, extensive empirical results demonstrate that our proposed MF-CSO not only shows clear advantages in terms of error rate reduction and dimensionality reduction but also has low computational costs compared to other EA-based FS

TABLE VI: Comparisons Between MF-CSO And Its Three Variants

| Dataset | Algorithm | Size | Error | S | Dataset | Algorithm | Size | Error | S |
|---|---|---|---|---|---|---|---|---|---|
| warpPIE10P | MF-CSO | 103.93 | **0.82** |  | ALLAML | MF-CSO | 297.6 | **3.76** |  |
|  | EMT-noKT | 195.8 | 1.83 | ≈ |  | EMT-noKT | 274.7 | 8.67 | - |
|  | MF-CSO-R | 118.5 | 0.89 | ≈ |  | MF-CSO-R | 298.9 | 9.50 | - |
|  | MF-PSO | 68.7 | 1.50 | ≈ |  | MF-PSO | 199.6 | 11.17 | - |
| Lymphoma | MF-CSO | 53.48 | **1.87** |  | Nci9 | MF-CSO | 2354.4 | **46.00** |  |
|  | EMT-noKT | 63.4 | 4.83 | - |  | EMT-noKT | 3716.7 | 50.17 | - |
|  | MF-CSO-R | 87.7 | 1.93 | ≈ |  | MF-CSO-R | 2260.3 | 53.59 | - |
|  | MF-PSO | 13 | 5.00 | - |  | MF-PSO | 3244.9 | 51.69 | - |
| Nci | MF-CSO | 273.46 | **28.77** |  | Adenocarcinoma | MF-CSO | 523.31 | **34.44** |  |
|  | EMT-noKT | 510.3 | 30.96 | - |  | EMT-noKT | 1165.4 | 35.53 | ≈ |
|  | MF-CSO-R | 250.7 | 39.03 | - |  | MF-CSO-R | 338.2 | 38.22 | - |
|  | MF-PSO | 427.6 | 34.46 | - |  | MF-PSO | 1237.3 | 38.93 | - |
| Leukemia 1 | MF-CSO | 229.6 | **7.5** |  | orlraws10P | MF-CSO | 322.51 | 3.70 |  |
|  | EMT-noKT | 318.8 | 11.67 | - |  | EMT-noKT | 362 | 7.00 | - |
|  | MF-CSO-R | 305.8 | 10.00 | - |  | MF-CSO-R | 397.2 | **2.00** | ≈ |
|  | MF-PSO | 260.7 | 16.39 | - |  | MF-PSO | 571.1 | 4.00 | ≈ |
| DLBCL | MF-CSO | 96.5 | **7.5** |  | Brain Tumor 2 | MF-CSO | 194.5 | **25.42** |  |
|  | EMT-noKT | 110.2 | 9.17 | ≈ |  | EMT-noKT | 511.9 | 32.92 | - |
|  | MF-CSO-R | 99.6 | 15.17 | - |  | MF-CSO-R | 294.8 | 35.00 | - |
|  | MF-PSO | 179.8 | 13.5 | - |  | MF-PSO | 461.5 | 33.75 | - |
| 9Tumor | MF-CSO | 199.2 | 55.23 |  | Prostate | MF-CSO | 63.2 | **12.67** |  |
|  | EMT-noKT | 346.1 | **45.6** | + |  | EMT-noKT | 308.2 | 13.00 | ≈ |
|  | MF-CSO-R | 337.8 | 56.67 | ≈ |  | MF-CSO-R | 304 | 13.50 | ≈ |
|  | MF-PSO | 369.9 | 56.38 | ≈ |  | MF-PSO | 561.4 | 14.00 | ≈ |
| TOX_171 | MF-CSO | 1170.7 | **7.62** |  | Leukemia2 | MF-CSO | 373.4 | **1.67** |  |
|  | EMT-noKT | 2559.3 | 9.25 | ≈ |  | EMT-noKT | 430.2 | 11.11 | - |
|  | MF-CSO-R | 2408.1 | 7.75 | ≈ |  | MF-CSO-R | 291.4 | 8.33 | - |
|  | MF-PSO | 3408.6 | 7.62 | ≈ |  | MF-PSO | 268.7 | 11.11 | - |
| Brain Tumor 1 | MF-CSO | 194.5 | 24.33 |  | 11 Tumor | MF-CSO | 334.72 | **20.31** |  |
|  | EMT-noKT | 1057 | 24.84 | ≈ |  | EMT-noKT | 1514.7 | 25.79 | - |
|  | MF-CSO-R | 217.1 | 27.31 | - |  | MF-CSO-R | 1162.3 | 23.03 | - |
|  | MF-PSO | 276 | **23.64** | ≈ |  | MF-PSO | 1237.3 | 21.15 | ≈ |
| Prostate6033 | MF-CSO | 212.13 | **12.82** |  | Lung Cancer | MF-CSO | 498.1 | **12.44** |  |
|  | EMT-noKT | 371.2 | 13.67 | ≈ |  | EMT-noKT | 779.3 | 13.13 | ≈ |
|  | MF-CSO-R | 245.8 | 14.83 | - |  | MF-CSO-R | 1773.4 | 23.62 | - |
|  | MF-PSO | 493.2 | 13.83 | ≈ |  | MF-PSO | 1180.8 | 15.64 | - |

methods.

*F. Ablation Experiment and Analysis*

1) *The Effect of Knowledge Transfer in EMT:* To investigate the effect of knowledge transfer in EMT, a variant of our proposed MF-CSO without performing knowledge transfer among relevant tasks, namely, EMT-noKT, is used here for performance comparison. That is, the tasks in EMT-noKT evolved separately without sharing useful knowledge. Table VI provides the average error rates and the numbers of selected features, where the better results are shown in **boldface**. As seen in Table VI, MF-CSO achieves lower error rates than EMT-noKT in all datasets except the '9 Tumors' dataset. In particular, the error rate of MF-CSO for the 'Leukemia2' dataset is approximately 10% lower than that of EMT-noKT. On the other hand, considering the size of the selected features, MF-CSO selects fewer features than EMT-noKT for all 18 datasets. In summary, the comparisons between the proposed MF-CSO and EMT-noKT indicate the effectiveness of knowledge transfer by sharing useful knowledge among the relevant FS tasks.

2) *The Effect of CSO for Multitasking Optimization:* To investigate the effect of the modified CSO-based knowledge transfer method in our method, a variant of MF-CSO is designed for performance comparison by replacing CSO with PSO, namely, MF-PSO. Notably, the adopted PSO-based knowledge transfer method in MF-PSO is the same as that in PSO-EMT [13]. As seen from Table VI, MF-CSO shows lower error rates than MF-PSO in all adopted datasets except the 'Brain Tumor 1' dataset. On the other hand, MF-CSO also performs better than MF-PSO in terms of dimensionality reduction on 14 out of 18 datasets, except the 'warpPIE10P', 'Lymphoma', 'ALLAML' and 'Leukemia2' datasets. In particular, for the 'Brain Tumor 2' dataset, the number of features selected by MF-CSO is less than half of that selected by MF-PSO, and our error rate is reduced by approximately 10%. Therefore, the superior performance obtained by MF-CSO over MF-PSO demonstrates the effectiveness of our proposed CSO-based knowledge transfer method. That is, the modified CSO variant is more effective and suitable than PSO when processing high-dimensional datasets.

3) *The Effect of Aggregation Mechanism in Knowledge Transfer:* To investigate the effect of the aggregation mechanism in our CSO-based knowledge transfer method, MF-CSO-R, which randomly selects a winner particle from one of the other three relevant tasks rather than using the aggregated winner particle from three relevant tasks to perform knowledge transfer, is designed for performance comparison. As shown in Table VI, the error rates obtained by MF-CSO are much lower than those of MF-CSO-R for 17 out of 18 datasets, and the error rate of MF-CSO on the 'orlraws10P' dataset is similar to that of MF-CSO-R. On the other hand, considering the size of the selected features, MF-CSO also shows advantages over MF-CSO-R in 14 out of 18 datasets except the 'Nci', 'Nci9', 'Adenocarcinoma' and 'Leukemia2' datasets. In particular, for the 'Lung Cancer' dataset, MF-CSO selects only one-tenth of the features of MF-CSO-R, with a lower error rate of approximately 10%. Therefore, these comparison results between MF-CSO and MF-CSO-R validate the effectiveness of our aggregation mechanism during the knowledge transfer process, as it allows better exploration of correlations between several relevant tasks to generate more promising winner particles for each task.

## V. Conclusion and Future Work

This paper proposed an efficient EMT-based FS method by using multiple filtering methods and CSO, termed MF-CSO, which can effectively handle high-dimensional classification problems. The implementation of EMT in our method can improve the performance of solving high-dimensional classification problems. In this method, the proposed multiple task generation strategy using multiple filtering methods can construct more diversified relevant tasks and has been validated to improve the effect of knowledge transfer in EMT when processing various kinds of high-dimensional datasets with different characteristics. Moreover, an efficient multiple task optimization strategy using a modified CSO in our method has been validated as more suitable to perform knowledge transfer among the relevant FS tasks. With the above two strategies, solving the original FS task can be highly accelerated by transferring useful knowledge from other relevant FS tasks. A number of experimental results have been conducted to verify the effectiveness of our proposed MF-CSO in solving eighteen high-dimensional datasets with the number of features ranging from 2000 to 12000. Compared to several state-of-the-art FS algorithms and three traditional filtering methods, MF-CSO can not only obtain a feature subset of higher quality in terms of classification accuracy and number of selected features but also has low computational costs.

Although the experimental results demonstrate that EMT is a promising technique for high-dimensional classification, it is meaningful and interesting to explore a general EMT framework to solve high-dimensional feature selection problems. Here, we provide the following outlook on possible improvements to this framework in terms of filtering methods and optimizers.

- *Filtering method*: In addition to the filtering methods used in this paper, more filtering methods with different characteristics can be embedded in the EMT framework to construct diverse FS tasks, thus improving the overall performance of processing various high-dimensional datasets. We give some suggestions for effective filtering methods that are not limited to them, which can be considered in the framework, such as symmetrical uncertainty [54], the Fisher score [55], and some extensions of *Relief-F* (e.g., *RRelief-F* [56], *ReliefMSS* [57], etc.).
- *Optimizer*: In addition to the CSO and PSO mentioned in this paper, we provide suggestions for efficient natural-inspired optimizers with distinct properties that are not limited to them, which can be considered in the framework, such as the artificial immune algorithm [58], differential evolution algorithm [59], and ant colony algorithm [60].

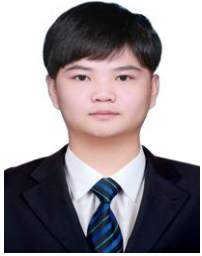

**Lingjie Li** received the B.S. degree from Shandong Technology and Business University, Yantai, China in 2017 and the M.S. degree from Shenzhen University, Shenzhen, China in 2020, where he is currently pursuing his Ph.D. in the College of Computer Science and Software Engineering.

He visited University of Exeter, Exeter, UK, as a visiting student in 2019. He focuses on research in the area of evolutionary computation, including algorithm studies on multi/many-objective optimization, large-scale optimization, and application studies on feature selection, cloud/edge computing, and remote sensing hyperspectral images.

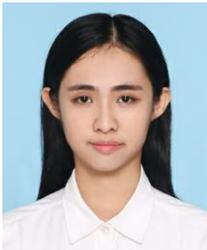

**Manlin Xuan** received the B.S. degree from Shandong University of Science and Technology, Qingdao, China in 2020. She is currently a master student in the College of Computer Science and Software Engineering, Shenzhen University.

Her current research interests include multi-objective optimization, evolutionary algorithms, evolutionary multitasking algorithms, and their applications to feature selection.

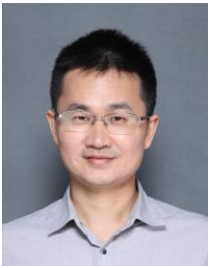

**Qiuzhen Lin** (Member IEEE) received the B.S. degree from Zhaoqing University and the M.S. degree from Shenzhen University, China, in 2007 and 2010, respectively. He received the Ph.D. degree from Department of Electronic Engineering, City University of Hong Kong, Kowloon, Hong Kong, in 2014.

He is currently an associate professor in College of Computer Science and Software Engineering, Shenzhen University. His current research interests include artificial immune system, multi-objective optimization, and dynamic system.

Prof. Lin has published over sixty research papers since 2008 and is currently serving as an Associate Editor for the IEEE TRANSACTIONS ON EVOLUTIONARY COMPUTATION.

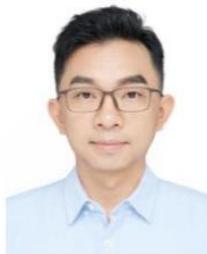

**Min Jiang** (Senior Member, IEEE) received the bachelor's and Ph.D. degrees in computer science from Wuhan University, Wuhan, China, in 2001 and 2007, respectively. Subsequently, he is a Postdoctoral Researcher with the Department of Mathematics, Xiamen University, Xiamen, China, where he is currently a Professor with the Department of Artificial Intelligence. His main research interests are machine learning, computational intelligence, and robotics. He has a special interest in dynamic multiobjective optimization, transfer learning, software development, and in the basic theories of robotics.

Prof. Jiang received the Outstanding Reviewer Award from IEEE TRANSACTIONS ON CYBERNETICS in 2016. He is the Chair of the IEEE CIS Xiamen Chapter. He is currently serving as an Associate Editor for the IEEE TRANSACTIONS ON NEURAL NETWORKS AND LEARNING SYSTEMS and IEEE TRANSACTIONS ON COGNITIVE AND DEVELOPMENTAL SYSTEMS.

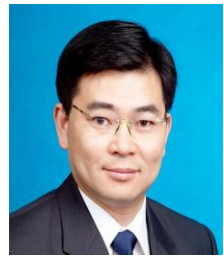

**Zhong Ming** received the Ph.D. degree in Computer Science and Technology from the Sun Yat-Sen University, Guangzhou, China, in 2004.

He is currently the Executive Director of the Graduate School of Shenzhen University, and a Professor with the National Engineering Laboratory for Big Data System Computing Technology and the College of Computer Science and Software Engineering, Shenzhen University, Shenzhen, China. His research interests include software engineering and artificial intelligence. He has published more than 200 refereed international conference and journal papers (including 40+ ACM/IEEE Transactions papers). He was the recipient of the ACM TiiS 2016 Best Paper Award and some other best paper awards.

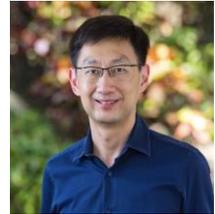

**Kay Chen Tan** (Fellow, IEEE) received the B.Eng. degree (First Class Hons.) and the Ph.D. degree from the University of Glasgow, U.K., in 1994 and 1997, respectively.

He is currently a Chair Professor (Computational Intelligence) of the Department of Computing, the Hong Kong Polytechnic University. He has published over 300 refereed articles and seven books.

Prof. Tan is currently the Vice-President (Publications) of IEEE Computational Intelligence Society, USA. He served as the Editor-in-Chief of the IEEE Computational Intelligence Magazine from 2010 to 2013 and the IEEE TRANSACTIONS ON EVOLUTIONARY COMPUTATION from 2015 to 2020. He currently serves as an Editorial Board Member for more than ten journals. Prof. Tan is an IEEE Distinguished Lecturer Program (DLP) Speaker and the Chief Co-Editor of Springer Book Series on Machine Learning: Foundations, Methodologies, and Applications.